\documentclass[%
 aip,
 amsmath,amssymb,
 reprint,%
]{revtex4-1}

\usepackage{graphicx}% Include figure files
\usepackage{dcolumn}% Align table columns on decimal point
\usepackage{bm}% bold math
\usepackage[utf8]{inputenc}
\usepackage[T1]{fontenc}
\usepackage{mathptmx}
\usepackage{etoolbox}
\usepackage{color}
\makeatletter
\def\@email#1#2{%
 \endgroup
 \patchcmd{\titleblock@produce}
  {\frontmatter@RRAPformat}
  {\frontmatter@RRAPformat{\produce@RRAP{*#1\href{mailto:#2}{#2}}}\frontmatter@RRAPformat}
  {}{}
}%
\makeatother
\begin{document}

\preprint{AIP/123-QED}

\title[Geometric Origins of Bias in Deep Neural Networks: A Human Visual System Perspective]{Geometric Origins of Bias in Deep Neural Networks: A Human Visual System Perspective}
% Force line breaks with \\
\author{Yanbiao Ma}
\email{ybma1998xidian@gmail.com, az381@cantab.ac.uk}
 %\altaffiliation[Also at ]{School of Artificial Intelligence, Xidian University, Xi’an, 710071, China}%Lines break automatically or can be forced with \\
\affiliation{ 
Gaoling School of Artificial Intelligence, Renmin University of China, Beijing, China%\\This line break forced with \textbackslash\textbackslash
}%

\author{Bowei Liu}%
\affiliation{ 
Tsinghua University, Beijing, China%\\This line break forced with \textbackslash\textbackslash
}%

\author{Andi Zhang*}%
\affiliation{ 
University of Manchester, Manchester, M13 9PS, United Kingdom%\\This line break forced with \textbackslash\textbackslash
}%

\date{\today}% It is always \today, today,
             %  but any date may be explicitly specified

\begin{abstract}
Bias formation in deep neural networks (DNNs) remains a critical yet poorly understood challenge, influencing both fairness and reliability in artificial intelligence systems. Inspired by the human visual system, which decouples object manifolds through hierarchical processing to achieve object recognition, we propose a geometric analysis framework linking the geometric complexity of class-specific perceptual manifolds in DNNs to model bias. Our findings reveal that differences in geometric complexity can lead to varying recognition capabilities across categories, introducing biases. To support this analysis, we present the Perceptual-Manifold-Geometry library, designed for calculating the geometric properties of perceptual manifolds. The toolkit has been downloaded and installed over 4,500 times. This work provides a novel geometric perspective on bias formation in modern learning systems and lays a theoretical foundation for developing more equitable and robust artificial intelligence.
\end{abstract}

\maketitle

\iffalse
\begin{quotation}
Bias formation in deep neural networks (DNNs) remains a critical yet poorly understood challenge, influencing both fairness and reliability in artificial intelligence systems. While previous studies have primarily focused on statistical imbalances in training data, recent advances suggest that the intrinsic geometry of feature representations may play a fundamental role in shaping classification biases. Motivated by findings in cognitive science and geometric deep learning, this study introduces a theoretical framework to analyze how perceptual manifold geometry affects class-wise recognition performance. By quantifying manifold complexity through curvature, dimensionality, and topological measures, this work reveals a strong correlation between geometric structure and classification bias. These findings contribute to a deeper theoretical foundation for understanding chaos and complexity in modern learning systems, offering new perspectives at the intersection of machine learning, neuroscience, and nonlinear dynamics.
\end{quotation}
\fi

%\medskip


\section{Introduction}

Deep neural networks (DNNs), with their powerful learning and generalization capabilities, have been widely applied in visual tasks such as image classification and object detection \cite{paper1,paper2}. However, the biases exhibited by DNNs toward different categories significantly limit their fairness and reliability in real-world applications \cite{paper5,paper6}. Traditional theories mainly attribute these biases to the long-tailed distribution of training samples \cite{paper10,paper11}. Nonetheless, research and practical observations suggest that even with balanced datasets, DNNs still show substantially better recognition performance for certain categories over others \cite{paper15,paper17}. This indicates that the mechanisms underlying such biases are more complex and remain poorly understood.

The human visual system provides critical insights into understanding the origins of biases in DNNs. When neurons in the visual cortex are stimulated by different physical attributes of objects belonging to the same category, they form Object manifolds \cite{paper20,paper21}. As shown in Figure.\ref{fig1}a, the human visual cortex gradually disentangles and reduces the dimensionality of complex Object manifolds layer by layer, making them easier to distinguish in deeper cortical areas, thereby achieving object recognition \cite{paper21,paper25}. This process suggests that the geometric complexity of Object manifolds influences the difficulty of disentanglement and recognition performance \cite{paper21,paper26,chaos1}.

Considering that the architecture of DNNs mimics this multi-layer disentanglement mechanism \cite{paper34,chaos2}, and recent studies \cite{paper19,paper29,paper31} demonstrate that the responses of DNNs to images exhibit manifold-like properties similar to those of the human visual system, we refer to the point-cloud manifolds formed by the embeddings of data in the feature space of a trained DNN's representation network as class-specific perceptual manifolds.
Formally, given a dataset \( X = [x_1, \dots, x_m] \) belonging to a specific class and a trained deep neural network \( \text{Model} = \{f(x, \theta_1), g(z, \theta_2)\} \), where \( f(x, \theta_1) \) represents the representation network and \( g(z, \theta_2) \) denotes the classifier, we extract the \( p \)-dimensional embeddings \( Z = [z_1, \dots, z_m] \in \mathbb{R}^{p \times m} \) using the representation network, with each \( z_i = f(x_i, \theta_1) \in \mathbb{R}^p \). The point-cloud manifold formed by \( Z \) is referred to as the \emph{class-specific perceptual manifold} within the DNN.  
If the recognition process of DNNs can be viewed as the representation network disentangling, reducing the dimensionality of, and separating class-specific perceptual manifolds, followed by the classifier making decisions (as illustrated in Figure.\ref{fig1}b), we hypothesize that: 
(1) The higher the geometric complexity of a class-specific perceptual manifold produced by the representation network, the more challenging it becomes for the classifier to decode and recognize the corresponding class. 
(2) Differences in geometric complexity across classes lead to inconsistent recognition capabilities, thus introducing biases. 

\section*{Results}
\begin{figure}[t]
\centering
\includegraphics[width=1\linewidth]{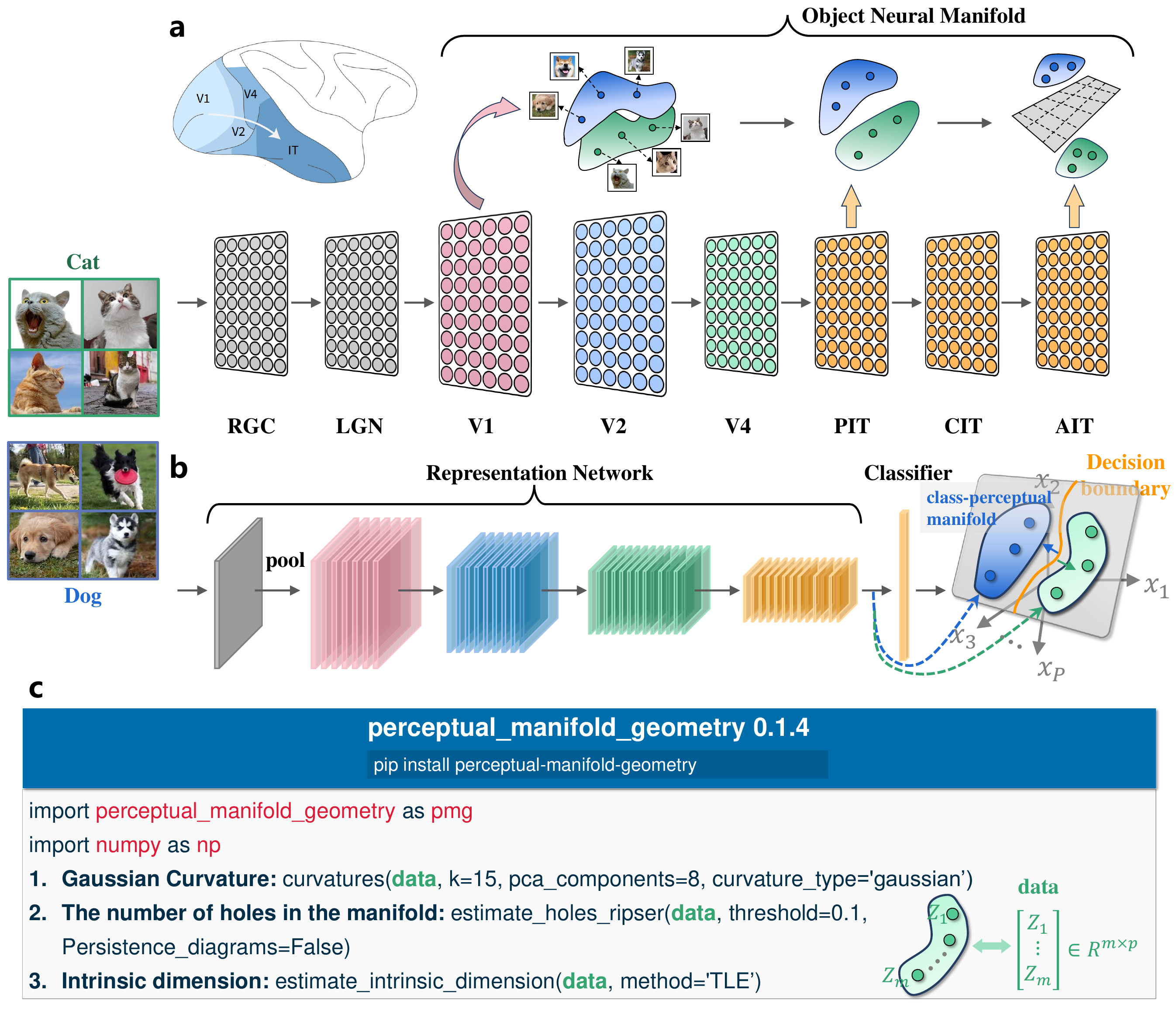}
\caption{\textbf{a}, When stimulated by objects of the same category, the human visual system forms object manifolds. As the visual cortex progresses through its layers, the object manifolds of different objects become more distinguishable and flatter.
\textbf{b}, Deep neural networks use representation networks to disentangle and reduce the dimensionality of data manifolds into class-specific perceptual manifolds, which are then used for object recognition by the classifier.
\textbf{c}, Example calculations of Gaussian curvature, the number of topological holes, and intrinsic dimensionality for perceptual manifolds using the \emph{Perceptual-Manifold-Geometry} toolkit.}
\label{fig1}
\end{figure}

To validate this hypothesis, we conducted experiments on widely used image datasets with balanced class samples, including CIFAR-10, CIFAR-100 \cite{cifar10}, Mini-ImageNet \cite{imagenet}, and Caltech-101 \cite{caltech}, to eliminate the influence of sample quantity. We employed deep neural networks with convolutional architectures, such as the ResNet \cite{resnet} series, and Transformer-based architectures, including ViT-B/16 and ViT-B/32 \cite{vit}. All models were trained following standard configurations (see Table 1) \cite{resnet,vit}. Each model was trained $10$ times using different random seeds, and the experimental results are reported as the average performance across these independent runs.
To systematically and efficiently quantify the geometric complexity of class-specific perceptual manifolds in DNNs, we developed a Python toolkit named \emph{Perceptual-Manifold-Geometry}. This toolkit offers comprehensive functionality for geometric analysis, including intrinsic dimensionality estimation, curvature analysis, and topological characteristics (the number of holes) quantification. Figure.\ref{fig1}c provides an example of the toolkit’s usage, see the Methods section for the relevant theory. Detailed documentation, tutorials, example datasets, and contribution guidelines are available online at \url{https://pypi.org/project/perceptual-manifold-geometry/}.

\begin{table}
\caption{\label{tab:table1}Summary of the experiments and results. LR denotes the learning rate and Mn denotes the momentum of the optimizer. If not specifically identified, the details of the experimental hyperparameters are shared in the Settings column. \textcolor[RGB]{0,201,87}{Green} and \textcolor{red}{red text} denote the hyperparameters of the convolutional neural network and vision transformer, respectively.}
\begin{ruledtabular}
\begin{tabular}{lll}
 \textbf{\normalsize Dataset} & \textbf{\normalsize DNN} & \textbf{\normalsize Settings} \\

CIFAR-10 & ViT-B/32, ViT-B/16  &  epoch: \textcolor[RGB]{0,201,87}{200}, \textcolor{red}{300} \\
CIFAR-100 & VGG-19  & Optimizer: SGD  \\
Mini-ImageNet  & SeNet-50   & Mm: 0.9 \\
Caltech-101  & ResNet-18, ResNet-34, ResNet-50  & LR: \textcolor[RGB]{0,201,87}{0.05}, \textcolor{red}{0.001} \\
\end{tabular}
\end{ruledtabular}
\end{table}

\begin{figure}[!t]
\centering
\includegraphics[width=1\linewidth]{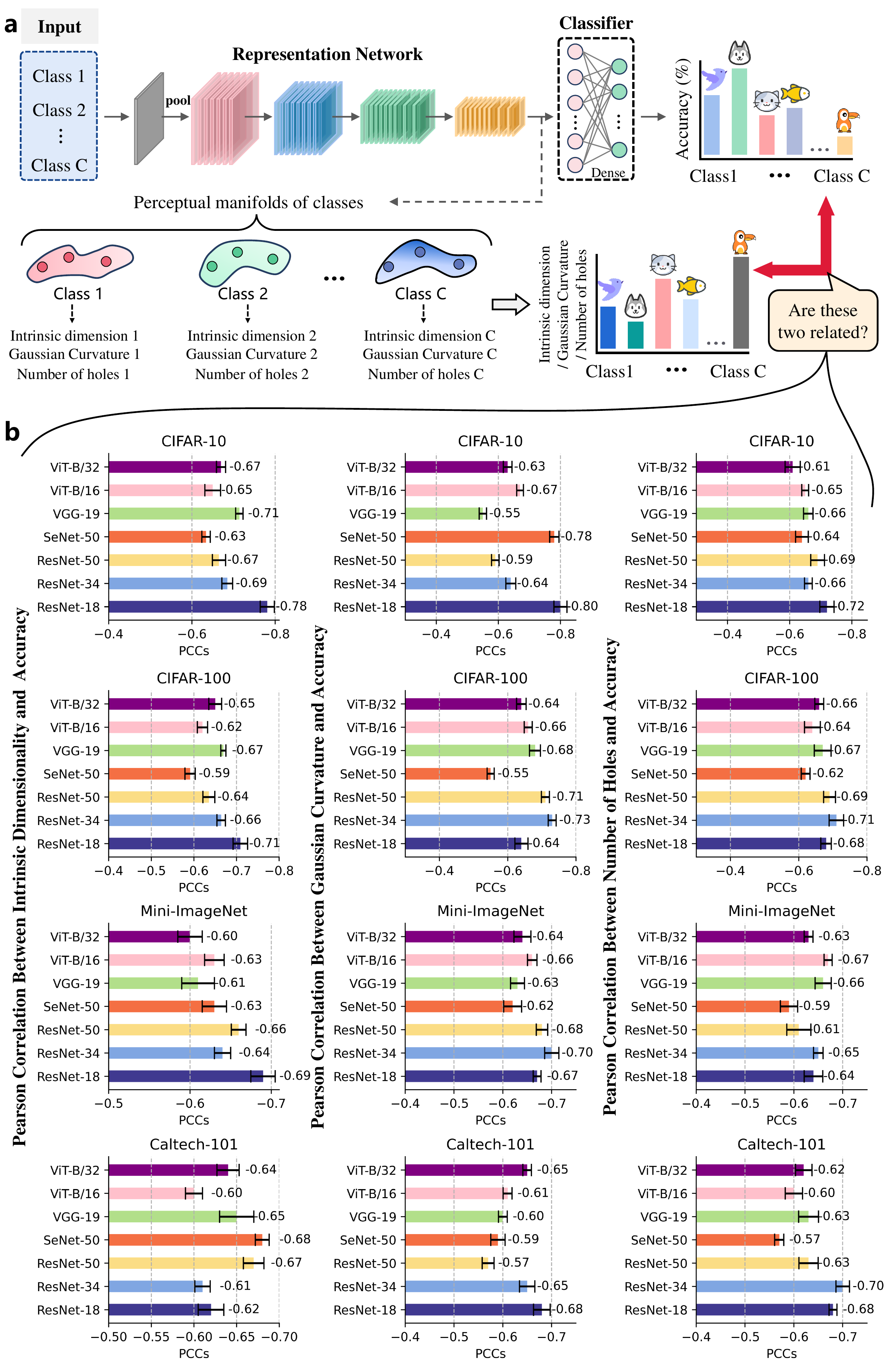}
\caption{\textbf{a}, Using the representation network of a DNN to extract image embeddings for each class and calculating the geometric complexity of class-specific perceptual manifolds based on these embeddings.
\textbf{b}, Bar charts showing Pearson correlation coefficients (PCCs) between class accuracy and various geometric complexities of class-specific perceptual manifolds across different datasets.}
\label{fig2}
\end{figure}

We first measured the recognition accuracy of fully trained DNNs on each class. As shown in Figure.\ref{fig2}a, we then extracted the embeddings of images for each class from the representation network of the DNN and stored them separately. Subsequently, we estimated the geometric complexity of the perceptual manifolds corresponding to each class’s embeddings, including intrinsic dimensionality, Gaussian curvature, and the number of topological holes. We calculated the Pearson correlation coefficients between geometric complexity and class-wise recognition accuracy.
As illustrated in Figure.\ref{fig2}b, the experimental results on different datasets and DNN architectures reveal a significant negative correlation between the geometric complexity of class-specific perceptual manifolds and recognition accuracy. These findings confirm our hypothesis: differences in geometric complexity among class-specific perceptual manifolds contribute to inconsistent recognition performance across classes, thereby inducing bias. Furthermore, the higher the geometric complexity of a perceptual manifold, the more challenging it is for the model to recognize the corresponding class.

This finding can also be understood from the perspective of model optimization. Intrinsic dimensionality reflects the complexity of a manifold's embedding. Higher dimensionality indicates more intricate manifold structures in high-dimensional space, requiring a classifier with greater capacity to effectively distinguish these samples. High Gaussian curvature typically signifies that the data distribution in the embedding space is more twisted and complex, leading to unstable decision boundaries and increasing classification difficulty. A higher number of topological holes implies more complex decision boundaries, making classifiers more prone to overfitting and resulting in poorer generalization performance.  
Our findings also offer new insights into mitigating model bias. By constraining optimization objectives, models can be encouraged to learn class-specific perceptual manifolds with lower and more balanced geometric complexity.  

Inspired by human visual behavior, this study establishes a universal geometric analysis framework to explain the pervasive bias in DNNs—an achievement that was previously difficult to realize through algorithmic research alone. Experimental results demonstrate that DNNs not only structurally resemble the human visual system but also exhibit similar characteristics in their internal data compression and processing mechanisms. This discovery reinforces the confidence of researchers in brain-inspired artificial intelligence.  
As the \emph{Perceptual-Manifold-Geometry} toolkit gains broader adoption, it is poised to provide valuable tools for artificial intelligence and neuroscience research. This work exemplifies the successful integration of neuroscience and computer science, highlighting the immense potential of interdisciplinary collaboration.

\section*{Methods}
To facilitate the study of perceptual manifolds in DNNs, we developed the Perceptual-Manifold-Geometry toolkit, which provides a comprehensive framework for analyzing the geometric characteristics of class-specific perceptual manifolds. The toolkit includes modules for intrinsic dimensionality estimation, Gaussian curvature calculation, and topological quantification, focusing particularly on the number of topological holes. These features enable systematic analysis of the relationships between manifold geometry and recognition performance.

\subsection*{Estimation of the Intrinsic Dimension}

Given a set of embeddings $Z=[z_1,\dots,z_n]\in \mathbb{R}^{p \times n}$ corresponding to an image dataset, $Z$ is typically distributed near a low-dimensional perceptual manifold $M$ embedded in the $p$-dimensional space, akin to a two-dimensional plane in three-dimensional space. The intrinsic dimension $ID(M)$ of the perceptual manifold is such that $d<p$. A higher intrinsic dimension indicates a more complex perceptual manifold. The following describes how to use TLE to estimate the intrinsic dimension of the perceptual manifold formed by $Z=[z_1,\dots,z_n]\in \mathbb{R}^{p \times n}$.

The primary method for estimating intrinsic dimension involves analyzing the distribution of distances between each point in the dataset and its neighboring points, and then estimating the dimensionality of the local space based on the rate of growth of distances or other statistic. Assuming that the distribution of samples is uniform within a small neighborhood, and then uses a Poisson process to simulate the number of points discovered by random sampling within neighborhoods of a given radius around each sample \cite{ID_MLE}. Subsequently, by constructing a likelihood function, the rate of growth in quantity is associated with the surface area of a sphere. Given any embedding $z_i$ in the dataset and its set of $k$ nearest neighbors $V$, the Maximum Likelihood Estimator (MLE) of the local intrinsic dimensiona at $z_i$ is given by: $$ID_{MLE}(z_i)=-\Bigg(\frac{1}{k}\sum_{j=1}^{k}ln\frac{r_j(z_i)}{r_k(z_i)}\Bigg)^{-1},$$ where $r_j(z_i)$ represents the distance between $z_i$ and its $j$-th nearest neighbor. TLE \cite{ID_TLE} no longer assumes uniformity of sample distribution in local neighborhoods, thus closer to the true data distribution. It assumes the local intrinsic dimension to be continuous, thereby utilizing nearby sample points to stabilize estimates at $z_i$. Specifically, the estimate of the intrinsic dimension at $z_i$ using TLE is given by 
\begin{equation}
ID_{TLE}(z_i) = -\Bigg(\frac{1}{\left | V_* \right |^2}\sum_{\substack{v,w\in V_*\\ v\neq w}}\Bigg[\ln\frac{d_{z_i}(v,w)}{r_k(z_i)} 
+\ln\frac{d_{z_i}(2z_i-v,w)}{r_k(z_i)}\Bigg]\Bigg)^{-1},
\end{equation}
where $V_* = V \cup \{z_i\}$, and $d_{z_i}(v,w)$ is defined as $(r_k(z_i)(w-v) \cdot (w-v)) / (2(z_i-v) \cdot (w-v))$. Furthermore, the global intrinsic dimensionality of the perceptual manifold is estimated as the average of local intrinsic dimensionalities: $$ID_{TLE}(Z)=\frac{1}{n}\sum_{i=1}^{n}ID_{TLE}(z_i).$$

\subsection*{Estimation of the Gaussian curvature}
Given a point cloud perceptual manifold $M$, which consists of a $p$-dimensional point set $Z=[z_1,\dots,z_n]\in \mathbb{R}^{p \times n}$, our goal is to calculate the Gauss curvature at each point. First, the normal vector at each point on $M$ is estimated by the neighbor points. Denote by $z_i^j$ the $j$-th neighbor point of $z_i$ and $u_i$ the normal vector at $z_i$. We solve for the normal vector by minimizing the inner product of $z_i^j-c_i,j=1,\dots,k$ and $u_i$ \cite{asao2021curvature}, i.e., $$\min {\textstyle \sum_{j=1}^{k}}((z_i^j-c_i)^Tu_i)^2,$$ where $c_i=\frac{1}{k}{\textstyle \sum_{j=1}^{k}}z_i^j$ and $k$ is the number of neighbor points. Let $y_j=z_i^j-c_i$, then the optimization objective is converted to
$$\min{\textstyle \sum_{j=1}^{k}}(y_j^Tu_i)^2 =\min {\textstyle \sum_{j=1}^{k}}u_i^Ty_jy_j^Tu_i =\min(u_i^T( {\textstyle \sum_{j=1}^{k}}y_jy_j^T)u_i).$$
${\textstyle \sum_{j=1}^{k}}y_jy_j^T$ is the covariance matrix of $k$ neighbors of $z_i$. Therefore, let $Y=[y_1,\dots,y_k]\in \mathbb{R}^{p\times k}$ and ${\textstyle \sum_{j=1}^{k}}y_jy_j^T=YY^T$. The optimization objective is further equated to
$$\begin{cases} f(u_i)=u_i^TYY^Tu_i,YY^T\in \mathbb{R}^{p\times p},
 \\ min(f(u_i)),
 \\ s.t. u_i^Tu_i=1.
\end{cases} $$
Construct the Lagrangian function $L(u_i,\lambda)=f(u_i)-\lambda (u_i^Tu_i-1)$ for the above optimization objective, where $\lambda$ is a parameter. The first-order partial derivatives of $L(u_i,\lambda)$ with respect to $u_i$ and $\lambda$ are
$$\frac{\partial L(u_i,\lambda)}{\partial u_i} =\frac{\partial}{\partial u_i}f(u_i)-\lambda\frac{\partial}{\partial u_i}(u_i^Tu_i-1) 
 =2(YY^Tu_i-\lambda u_i), $$ 
$$\frac{\partial L(u_i,\lambda)}{\partial \lambda}  =u_i^Tu_i-1.$$
Let $\frac{\partial L(u_i,\lambda)}{\partial u_i}$ and $\frac{\partial L(u_i,\lambda)}{\partial \lambda}$ be $0$, we can get $YY^Tu_i=\lambda u_i,u_i^Tu_i=1$. It is obvious that solving for $u_i$ is equivalent to calculating the eigenvectors of the covariance matrix $YY^T$, but the eigenvectors are not unique. From $\left \langle YY^Tu_i,u_i \right \rangle =\left \langle \lambda u_i,u_i \right \rangle$ we can get $\lambda=\left \langle YY^Tu_i,u_i \right \rangle=u_i^TYY^Tu_i $, so the optimization problem is equated to $\mathop{\arg\min}_{u_i}(\lambda)$. Performing the eigenvalue decomposition on the matrix $YY^T$ yields $p$ eigenvalues $\lambda_1,\dots,\lambda_p$ and the corresponding $p$-dimensional eigenvectors $[\xi_1,\dots,\xi_p]\in \mathbb{R}^{p\times p}$, where $\lambda_1\ge \dots \ge \lambda_p\ge 0$, $\left \| \xi_i \right \| _2=1,i=1,\dots,p$, $\left \langle \xi_a,\xi_b \right \rangle =0(a\neq b)$. The eigenvector $\xi_{m+1}$ corresponding to the smallest non-zero eigenvalue of the matrix $YY^T$ is taken as the normal vector $u_i$ of $M$ at $z_i$.

Consider an $m$-dimensional affine space with center $z_i$, which is spanned by $\xi_1,\dots,\xi_m$. This affine space approximates the tangent space at $z_i$ on $M$. We estimate the curvature of $M$ at $z_i$ by fitting a quadratic hypersurface in the tangent space utilizing the neighbor points of $z_i$. The $k$ neighbors of $z_i$ are projected into the affine space $z_i+\left \langle \xi_1,\dots,\xi_m \right \rangle$ and denoted as 
$$o_j=[(z_i^j-z_i)\cdot \xi_1,\dots,(z_i^j-z_i)\cdot \xi_m]^T\in \mathbb{R}^m,j=1,\dots,k.$$
Denote by $o_j[m]$ the $m$-th component $(z_i^j-z_i)\cdot \xi_m$ of $o_j$. We use $z_i$ and $k$ neighbor points to fit a quadratic hypersurface $f(\theta)$ with parameter $\theta \in \mathbb{R}^{m\times m}$. The hypersurface equation is denoted as
$$f(o_j,\theta)=\frac{1}{2} {\textstyle \sum_{a,b}}\theta_{a,b}o_j\left [ a \right ] o_j\left [ b \right ] ,j\in \left \{ 1,\dots,k \right \}, $$
further, minimize the squared error
$$E(\theta)= {\textstyle \sum_{j=1}^{k}}( \frac{1}{2} {\textstyle \sum_{a,b}}\theta_{a,b}o_j\left [ a \right ] o_j\left [ b \right ] -(z_i^j-z_i)\cdot u_i)^2.$$
Let $\frac{\partial E(\theta)}{\partial \theta_{a,b}}=0,a,b\in \left \{ 1,\dots,m \right \}$ yield a nonlinear system of equations, but it needs to be solved iteratively. Here, we propose an ingenious method to fit the hypersurface and \textbf{give the analytic solution of the parameter $\theta$} directly. Expand the parameter $\theta$ of the hypersurface into the column vector
$$\theta=\left [ \theta_{1,1},\dots,\theta_{1,m},\theta_{2,1},\dots,\theta_{m,m} \right]^T\in \mathbb{R}^{m^2}.$$
Organize the $k$ neighbor points $\left \{ o_j \right \}_{j=1}^k $ of $z_i$ according to the following form:
$$ O(z_i)=\begin{bmatrix} o_1\left [ 1 \right ] o_1\left [ 1 \right ] 
  &  o_1\left [ 1 \right ] o_1\left [ 2 \right ]  & \cdots  &  o_1\left [ m \right ] o_1\left [ m \right ]  \\
  o_2\left [ 1 \right ] o_2\left [ 1 \right ]  & o_2\left [ 1 \right ] o_2\left [ 2 \right ] & \cdots &o_2\left [ m \right ] o_2\left [ m \right ] \\
 \vdots  & \vdots  & \ddots  & \vdots  \\
 o_k\left [ 1 \right ] o_k\left [ 1 \right] & o_k\left [ 1 \right ] o_k\left [ 2 \right] & \cdots  &o_k\left [ m \right ] o_k\left [ m \right]
\end{bmatrix}\in \mathbb{R}^{k\times m^2}.$$
The target value is
$$T=\left [ (z_i^1-z_i)\cdot u_i,(z_i^2-z_i)\cdot u_i,\dots,(z_i^k-z_i)\cdot u_i \right ]^T \in \mathbb{R}^{k}.$$
We minimize the squared error
$$E(\theta)=\frac{1}{2}tr\left [ \left ( O(z_i)\theta-T \right)^T(O(z_i)\theta-T) \right ],$$
and find the partial derivative of $E(\theta)$ for $\theta$:
$$\frac{\partial E(\theta)}{\partial \theta} =\frac{1}{2} \left ( \frac{\partial tr(\theta^TO(z_i)^TO(z_i)\theta)}{\partial \theta}-\frac{\partial tr(\theta^TO(z_i)^TT)}{\partial \theta}   \right )$$
$$=O(z_i)^TO(z_i)\theta-O(z_i)^TT.$$
Let $\frac{\partial E(\theta)}{\partial \theta}=0$, we can get
$$\theta = (O(z_i)^TO(z_i))^{-1}O(z_i)^TT.$$
Thus, the Gauss curvature of the perceptual manifold $M$ at $z_i$ can be calculated as
$$G(z_i)=det(\theta)=det((O(z_i)^TO(z_i))^{-1}O(z_i)^TT).$$

Up to this point, we provide an approximate solution of the Gauss curvature at any point on the point cloud perceptual manifold $M$. Recent research \cite{balestriero2021learning} shows that on a high-dimensional dataset, almost all samples lie on convex locations, and thus the complexity of the perceptual manifold is defined as the average $\frac{1}{n} {\textstyle \sum_{i=1}^{n}}G(z_i)$ of the Gauss curvatures at all points on $M$. Our approach does not require iterative optimization and can be quickly deployed in a deep neural network to calculate the Gauss curvature of the perceptual manifold.

\subsection*{Estimation of the Number of Holes}

To quantitatively characterize the topological complexity of class-specific perceptual manifolds, we employ \textbf{Persistent Homology (PH)}, a core tool in Topological Data Analysis (TDA) that captures multi-scale topological features—such as connected components, loops (1-dimensional holes), and voids (2-dimensional cavities)—in high-dimensional point cloud data. Unlike traditional geometric measures, PH provides a robust, coordinate-invariant description of shape structure, making it particularly suitable for analyzing the nonlinear embedding geometries learned by deep neural networks \cite{carlsson2009topology}.

Specifically, the Ripser algorithm is used to compute its Persistence Diagram, capturing the evolution of topological structures at different scales. The persistence diagram represents the topological complexity of the manifold by recording the birth and death times of each topological feature. In this study, we focus on 1-dimensional homology \( H_1 \), which corresponds to loop structures, where the total number of loops reflects the number of holes in the data manifold.

To enhance computational robustness, a persistence threshold \( \tau \) is introduced to filter out low-persistence features caused by noise. Let \( \text{Persistence} = \text{Death} - \text{Birth} \); only loops with \( \text{Persistence} > \tau \) are retained as significant features. The number of holes is then defined as the total count of these significant loops. Additionally, to characterize the spatial distribution of holes comprehensively, we compute their total persistence, average persistence, and persistence density (defined as the ratio of total persistence to the time span of features).

%This method, implemented efficiently using tools such as Ripser \cite{bauer2021ripser}, extracts key topological attributes of the data without significantly increasing computational complexity. These topological metrics not only reveal the geometric complexity of perceptual manifolds but also provide a theoretical foundation for understanding their influence on classifier decision boundaries.

Given a finite set of embeddings $ Z = \{z_1, z_2, \dots, z_n\} \subset \mathbb{R}^p $ corresponding to images of a specific class, we model the underlying perceptual manifold $ M $ as a geometric object whose topology can be inferred from the proximity relationships among points in $ Z $. The central idea of persistent homology is to construct a nested sequence of simplicial complexes—called a filtration—over $ Z $, parameterized by an increasing scale $ \epsilon \geq 0 $, and track how topological features emerge and disappear across scales.

\textbf{Filtration Construction: Vietoris–Rips Complex}
We adopt the \textbf{Vietoris–Rips (VR) filtration}, which builds a simplicial complex $ \mathrm{VR}(Z, \epsilon) $ at each scale $ \epsilon $ as follows:
- For $ \epsilon = 0 $, $ \mathrm{VR}(Z, 0) $ consists only of isolated vertices.
- As $ \epsilon $ increases, an edge is added between any two points $ z_i, z_j $ if $ \|z_i - z_j\| < \epsilon $.
- A $ k $-simplex (e.g., triangle for $ k=2 $) is included if all pairwise distances among its $ k+1 $ vertices are less than $ \epsilon $.

This yields a nested family of complexes:
$$
\mathrm{VR}(Z, \epsilon_1) \subseteq \mathrm{VR}(Z, \epsilon_2) \subseteq \cdots \subseteq \mathrm{VR}(Z, \epsilon_{\max}), \quad  
$$
$$
\text{for } \epsilon_1 < \epsilon_2 < \cdots < \epsilon_{\max}.
$$

\textbf{Homology Groups and Betti Numbers}
At each scale $ \epsilon $, we compute the \textbf{homology groups} $ H_k(\mathrm{VR}(Z, \epsilon)) $, which algebraically encode the number of independent $ k $-dimensional topological features:
- $ H_0 $: Connected components (clusters),
- $ H_1 $: One-dimensional loops (holes),
- $ H_2 $: Two-dimensional voids (cavities), etc.
The rank of $ H_k $, denoted $ \beta_k(\epsilon) $, is the \textbf{$ k $-th Betti number}, representing the number of linearly independent $ k $-dimensional holes at scale $ \epsilon $. In this work, we focus on $ \beta_1(\epsilon) $, which quantifies the number of non-contractible loops in the perceptual manifold—a direct measure of its topological complexity.
However, individual Betti numbers are sensitive to noise and depend heavily on the choice of $ \epsilon $. To overcome this limitation, persistent homology tracks the \emph{birth} and \emph{death} of each topological feature across the filtration.

\textbf{Persistence Diagram and Persistence Barcode}
Each topological feature (e.g., a loop) appears ("is born") at some scale $ \epsilon_b $ and disappears ("dies") at $ \epsilon_d > \epsilon_b $. This gives rise to a pair $ (\epsilon_b, \epsilon_d) $. The collection of all such pairs forms the persistence diagram $ Dgm_k(Z) \subset \mathbb{R}^2 $ for dimension $ k $. Alternatively, these intervals can be visualized as horizontal bars in a persistence barcode, where longer bars represent more persistent (i.e., structurally significant) features.

The persistence of a feature is defined as:
$$
\mathrm{Persistence} = \epsilon_d - \epsilon_b.
$$

Features with small persistence are typically artifacts of sampling noise or local perturbations, while those with large persistence reflect global, stable structures.

\textbf{Quantification of Topological Holes}
To obtain a robust scalar metric for comparison across classes and models, we define the number of significant topological holes as:
$$
N_{\text{holes}} = \# \left\{ (\epsilon_b, \epsilon_d) \in Dgm_1(Z) \mid \epsilon_d - \epsilon_b > \tau \right\},
$$
where $ \tau $ is a persistence threshold chosen to filter out noisy features (we set $ \tau = 0.1 \times \mathrm{diam}(Z) $ empirically, with $ \mathrm{diam}(Z) = \max_{i,j} \|z_i - z_j\| $).

In addition to counting holes, we compute complementary metrics:
\begin{itemize}
\item Total Persistence: $ P_{\text{total}} = \sum_{(\epsilon_b,\epsilon_d)} (\epsilon_d - \epsilon_b) $
\item Average Persistence: $ P_{\text{avg}} = \frac{P_{\text{total}}}{N_{\text{holes}}} $
\item Persistence Density: $ P_{\text{density}} = \frac{P_{\text{total}}}{\epsilon_{\max} - \epsilon_{\min}} $
\end{itemize}
These metrics provide a multi-faceted view of the topological structure, capturing both the quantity and stability of holes.

\textbf{Computational Implementation}
We implement this pipeline using the efficient \texttt{Ripser} library \cite{bauer2021ripser}, which computes sparse VR complexes and persistence diagrams in low memory and time complexity ($ O(n^2) $ to $ O(n^3) $ depending on sparsity). All computations are performed on GPU-accelerated hardware to enable large-scale analysis across datasets and architectures.

By integrating persistent homology into our geometric framework, we move beyond purely metric or differential geometric descriptions and capture intrinsic topological constraints that influence classifier decision boundaries. A higher number of persistent 1D holes implies more complex, entangled manifold structures, leading to fragmented or irregular decision regions and increased classification difficulty—consistent with our empirical findings.

%\section*{Limitations}
%While our geometric framework provides new insights into bias formation in DNNs, several limitations should be acknowledged. First, the current implementation of the Perceptual-Manifold-Geometry toolkit assumes that perceptual manifolds can be well-approximated by point clouds in Euclidean space, which may not fully capture non-Euclidean or hierarchical structures present in deeper layers of certain architectures (e.g., hyperbolic representations). Second, the estimation of Gaussian curvature and intrinsic dimensionality relies on local neighborhood assumptions, whose accuracy depends on the choice of $k$-nearest neighbors and embedding density. Although we conducted sensitivity analyses (see Supplementary Information), future work could explore adaptive neighborhood selection strategies. Third, while our analysis spans multiple datasets and architectures, it primarily focuses on image classification tasks; extension to other modalities (e.g., speech, text) remains an open direction.

\section*{Future Directions}
Our framework opens several promising avenues for future research. One direction is to integrate geometric regularization into training objectives—such as penalizing high-curvature or high-dimensional manifolds—to actively reduce bias during learning. Another is to extend this analysis to dynamic models (e.g., Transformers over time) to study how perceptual manifolds evolve during sequence processing. Furthermore, combining neuroimaging data with DNN manifold geometry could enable direct comparisons between artificial and biological vision systems, advancing brain-inspired AI.

\section*{Conflict of Interest Statement}
The author (authors) has (have) no conflicts to disclose.

\section*{Acknowledgements}
Andi Zhang contributed to the computational resources used in this study.

\section*{Author contributions}
Yanbiao Ma: Conceptualization, Data curation, Formal analysis, Investigation, Methodology, Software, Writing – original
draft. Bowei Liu: Conceptualization, Data curation, Formal
analysis, Investigation, Methodology, Software. Andi Zhang: Funding acquisition,
Validation. All authors have read and agreed to the published
version of the manuscript.

\section*{Data availability}
All deep neural networks used in this study have been appropriately cited in the main text. All datasets utilized in this research are open access and have been referenced in the paper. The web links to the datasets are as follows: CIFAR-10/100 (\url{https://www.cs.toronto.edu/~kriz/cifar.html}), Caltech-101 (\url{https://www.vision.caltech.edu/datasets/}). The Mini-ImageNet dataset is derived from ImageNet and can be generated using the MLclf package (\url{https://github.com/tiger2017/MLclf}). Additionally, the toolkit developed for calculating the geometric properties of perceptual manifolds has been officially released and is available at \url{https://github.com/mayanbiao1234/Geometric-metrics-for-perceptual-manifolds}.

\bibliography{aipsamp}% Produces the bibliography via BibTeX.

\end{document}